\documentclass[letterpaper, 10 pt, conference]{ieeeconf}  

\IEEEoverridecommandlockouts                              
\usepackage{amsmath,amsfonts}

\usepackage{bm}
\usepackage{algorithmic}
\usepackage{algorithm}

\usepackage[caption=false]{subfig}

\usepackage{array}
\usepackage{textcomp}
\usepackage{stfloats}
\usepackage{url}
\usepackage{verbatim}
\usepackage{graphicx}

\usepackage{booktabs}
\usepackage{tabularx}

\usepackage{cite}

\makeatletter
\let\NAT@parse\undefined
\makeatother

\usepackage[colorlinks,linkcolor=black,anchorcolor=black,citecolor=black,urlcolor=black,hyperfootnotes=true]{hyperref}   
\usepackage[all]{hypcap} 

\title{\LARGE \bf
Nonlinear MPC for Quadrotors in Close-Proximity Flight \\ with Neural Network Downwash Prediction
}

\author{Jinjie Li$^{1}$, Liang Han$^{2*}$, Haoyang Yu$^{2}$, Yuheng Lin$^{2}$, Qingdong Li$^{1}$, Zhang Ren$^{1}$
\thanks{This work was supported by the National Natural Science Foundation of China under Grants 62373024 and 61803014, the Zhejiang Provincial Natural Science Foundation of China under Grant LGG22F030025, the Aeronautical Science Foundation of China under Grant 2022Z071051015, the Science and Technology Innovation 2030-Key Project of ``New Generation Artiﬁcial Intelligence" under Grant 2018AAA0102305, and the Fundamental Research Funds for the Central Universities. We thank Dr. Sihao Sun for insightful discussions.}
\thanks{$^{1}$J. Li, Q. Li, and Z. Ren are with the School of Automation Science and Electrical Engineering, Beihang University, Beijing, 100191, China
{\tt\small \{lijinjie, liqingdong, renzhang\}@buaa.edu.cn}}%
\thanks{$^{2}$L. Han (*Corresponding author), H. Yu, Y. Lin are with the Sino-French Engineer School, Beihang University, Beijing, 100191, China
{\tt\small \{liang\_han, haoyang\_yu, linyuheng\}@buaa.edu.cn}}%
}

\begin{document}

\maketitle
\thispagestyle{empty}
\pagestyle{empty}

\begin{abstract}

Swarm aerial robots are required to maintain close proximity to successfully traverse narrow areas in cluttered environments. However, this movement is affected by the downwash effect generated from other quadrotors in the swarm. This aerodynamic effect is highly nonlinear and hard to describe through mathematical modeling. Additionally, the existence of the downwash disturbance can be predicted based on the states of neighboring quadrotors. If this prediction is considered, the control loop can proactively handle the disturbance, resulting in improved performance.

To address these challenges, we propose an approach that integrates a Neural network Downwash Predictor with Nonlinear Model Predictive Control (NDP-NMPC). The neural network is trained with spectral normalization to ensure robustness and safety in uncollected cases. The predicted disturbances are then incorporated into the optimization scheme in NMPC, which enforces constraints to ensure that states and inputs remain within safe limits. We also design a quadrotor system, identify its parameters, and implement the proposed method on board. Finally, we conduct a prediction experiment to validate the safety and effectiveness of the network. In addition, a real-time trajectory tracking experiment is performed with the entire system, demonstrating a 75.37\% reduction in tracking error in height under the downwash effect.


\end{abstract}

\section*{Supplementary Material}

We make the code and dataset available to the community at: \href{https://github.com/Li-Jinjie/ndp_nmpc_qd}{{\tt\small  https://github.com/Li-Jinjie/ndp\_nmpc\_qd}}.

\section{Introduction}

\begin{figure}[t]
    \centerline{\includegraphics[trim=0 0 0 0,clip,width=\linewidth]{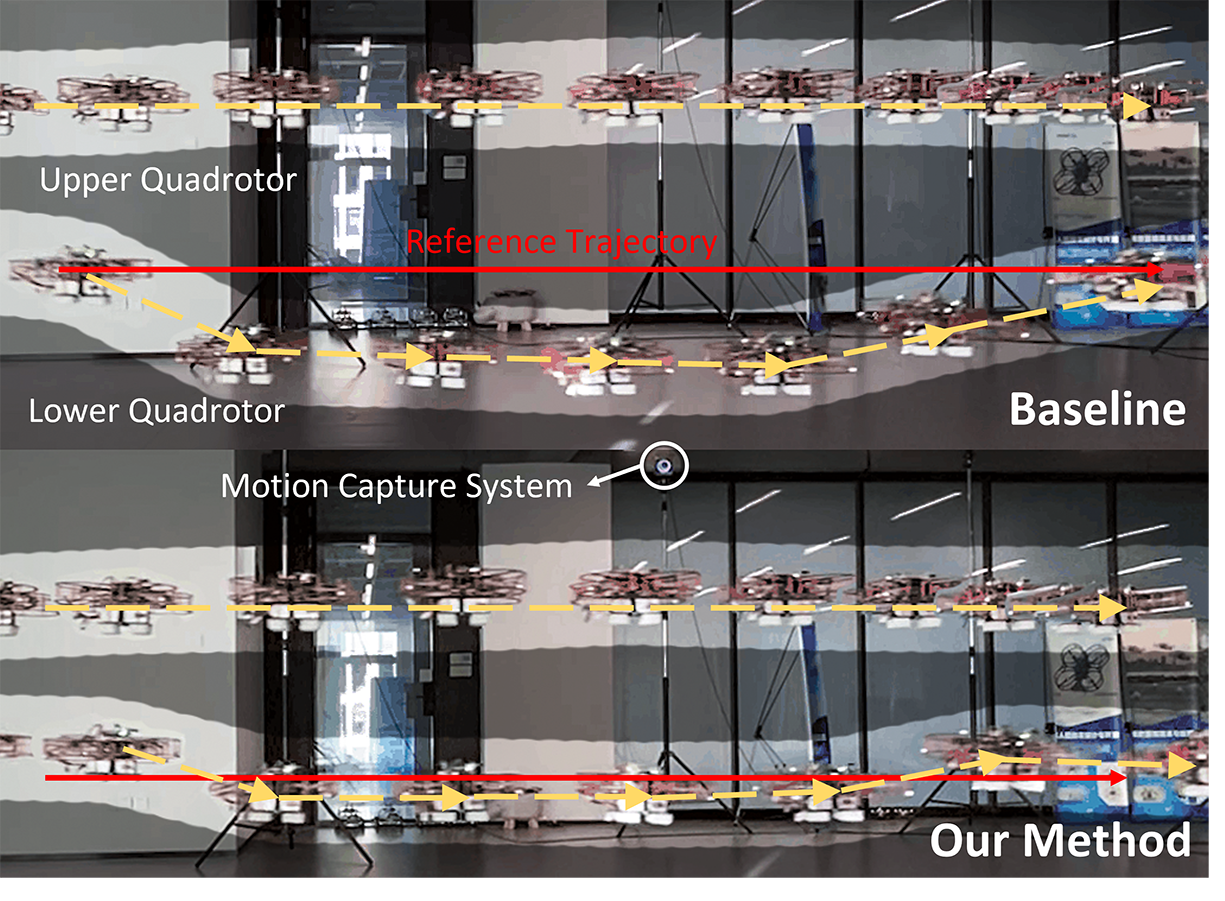}} 
    \caption{Comparison of two methods in real-world close-proximity flight. In each scenario, two quadrotors fly together to track reference trajectories from left to right, with the lower quadrotor experiencing downwash effects from the upper drone.
    }
    \label{fig:real_comparison}
\end{figure}

Advances in swarm robots have attracted significant attention since they can utilize cooperation and coordination among a group of robots \cite{zhou_swarm_2022} to achieve complex tasks that are impossible for a single robot. As one type of swarm robot, aerial swarm robots differ from ground robots in that each agent can be affected by the strong airflow generated from its upper neighbors, which is referred to as the downwash effect \cite{chung_survey_2018}. When an agent cannot accurately track its trajectory due to the airflow disturbances, it could influence other agents in normal flight (Fig. \ref{fig:real_comparison}), thereby amplifying the disturbance until it affects the entire system. Thus, addressing the downwash effect is crucial for ensuring the flight safety of aerial swarm robots.

As the distances among aerial robots vary, the downwash effect exhibits a corresponding variation in strength, with greater intensity observed at shorter distances and weaker effects observed at longer distances. Hence, some researches \cite{honig_trajectory_2018, arul_dcad_2020} treat the quadrotor as an ellipsoid with a long vertical axis, attempting to avoid close vertical flight when planning trajectories. However, this assumption reduces the \textit{reachable set} of the aerial swarm robots, preventing pushing their mobility to the boundary. A better approach is to consider the downwash effect as a \textit{disturbance rejection} problem and address it within the control layer \cite{shi_neural-swarm2_2021}. In this way, the planning layer only needs to consider the collision radius of each agent and thus maximizes mobility. Therefore, it is necessary to model the downwash disturbance and incorporate it into the control loop.

\setlength{\textfloatsep}{8pt plus 1.0pt minus 2.0pt}


The downwash effect is difficult to model by mathematical equations due to its high-degree nonlinearity. Traditionally, the airflow effect can be accurately modeled through real-world experiments \cite{yeo_empirical_2015, carter_influence_2021} or Computational Fluid Dynamics (CFD) simulation \cite{jimenez-cano_contact-based_2019}, while these approaches all pose high requirements for experimental equipment or computational time. On the other hand, the rapid development of deep learning in recent years renders the possibility to simulate nonlinear phenomena such as airflow disturbance in low time and fund demands \cite{willard_integrating_2022}. Shi et al. conduct a series of pioneer research to apply deep learning to model the airflow phenomena in aerial robotics, including the ground effect when the quadrotor lands \cite{shi_neural_2019}, the downwash disturbances among quadrotors \cite{shi_neural-swarm2_2021}, and the disturbances in strong winds \cite{oconnell_neural-fly_2022}, which demonstrate the feasibility of modeling downwash effect using neural networks (NNs).

When integrating the NN model inside the control loop, the research \cite{shi_neural-swarm2_2021} adopts a hierarchical feedback-linearization controller, which generates control inputs based on only the current states. However, the downwash effect is generated by the relative motions of other quadrotors and hence can be predicted through the exchange of reference trajectories. If the controller can fully exploit this prediction, the overshoots when tracking trajectories will be reduced, and thus the tracking performance is improved. As a prediction-based method, Model Predictive Control (MPC) can look forward and is a suitable solution to integrate the downwash prediction. Besides, MPC has the advantage of handling constraints \cite{sun_comparative_2022}, which can avoid the input saturation of quadrotors when resisting disturbances.

Numerous studies have attempted to combine deep networks with MPC and to assess their performance on a single quadrotor \cite{salzmann_real-time_2023, chee_knode-mpc_2022, bauersfeld_neurobem_2021}, 
while limited research tries to apply this combination to the downwash problem. Matei et al. \cite{matei_controlling_2021} apply an MPC-based controller with a learning-driven interaction model to solve the downwash problem. Nevertheless, their approach uses a network as the MPC dynamics to accelerate computation, which is less accurate than a physics-based model and cannot be run onboard in real-time.

In this work, we design a trajectory tracking system for close-proximity flight by integrating a neural network disturbance predictor with Nonlinear Model Predictive Control (NMPC). The proposed approach is inspired by Shi et al. \cite{shi_neural-swarm2_2021} and extends it to NMPC to fully exploit predictive power. First, using motor speed sensors and a physical model to collect downwash data, we train a Multi-Layer Perceptron (MLP) to predict disturbances and utilize \textit{spectral normalization} to ensure robustness. Then, we integrate the predictor with NMPC to propose the trajectory tracking method. We also briefly introduce the trajectory generation algorithm to close the loop. Finally, we implement the proposed approach on two quadrotors to verify its effectiveness.


\setlength{\textfloatsep}{8pt plus 1.0pt minus 2.0pt}  
\begin{figure*}[t]
    \centerline{\includegraphics[trim=0 5 0 0,clip,width=\textwidth]{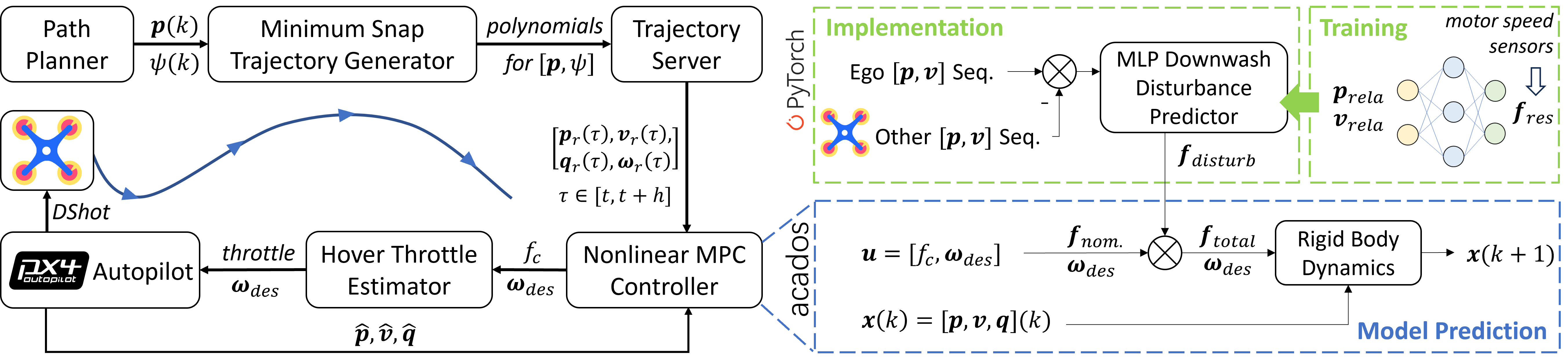}} 

    \vspace*{-1mm}
    \caption{The overall workflow of the proposed method, which follows the \textit{Path Planning}-\textit{Trajectory Generation}-\textit{Control} pipeline. Specifically, Nonlinear Model Predictive Control (NMPC) is adopted for trajectory tracking, of which the model is a nominal quadrotor model assisted by a predicted disturbance sequence calculated before every iteration. These disturbances are predicted by a network using the error state sequence of its own and other quadrotors.
    }
    \label{fig:workflow}
\end{figure*}

The main contributions are as follows:
\begin{enumerate}
\item an NMPC-based trajectory tracking method with network disturbance prediction (NDP-NMPC) to exploit predicted movement information and to address the saturation constraints under close-proximity flight,
\item real-time experiments to verify trajectory tracking performance under downwash effects, and
\item provision of open-source code and a dataset to support further research in this area.
\end{enumerate}


\section{Methodology}
This section outlines our proposed control scheme that combines a neural network disturbance observer with NMPC to enhance tracking performance during close-proximity flight. The overall workflow is presented in Section III-A. Then, subsections B and C introduce notation, coordinate systems, and a nominal quadrotor model. Leveraging these conventions, a neural network observer is implemented to predict the downwash disturbance in Section III-D. Finally, a modified NMPC trajectory tracking controller with disturbance prediction is proposed in Section III-E. This subsection also briefly introduces the generation of reference trajectories, which is essential to practical implementation.

\subsection{System Overview}

The system architecture is illustrated in Fig. \ref{fig:workflow}. Moving from left to right, a sequence of position points and yaw angles is initially transmitted from a \textit{Path Planner} to a \textit{Trajectory Generator}. The latter module then leverages the minimum snap method \cite{mellinger_minimum_2011} to generate a continuous polynomial trajectory from multiple derivatives of position and yaw angle. Subsequently, a \textit{Trajectory Server} discretizes the trajectory and computes the desired full states through differential flatness \cite{mellinger_minimum_2011}. These full-state points are transmitted as control reference to an \textit{NMPC Controller} at a high frequency for computing control outputs. The control command, after being converted from force to throttle by a \textit{Hover Throttle Estimator}, is ultimately executed by the \textit{PX4 Autopilot} \cite{meier_px4_2015} to operate the quadrotor. The estimated states are fed back from the \textit{Autopilot} to the \textit{Controller} for closing the loop.


The downwash effect is taken into account within the NMPC controller. NMPC is a model-based control approach, and its model consists of two components: one based on a nominal quadrotor model and the other based on a neural network disturbance predictor. Prior to flight, the network is trained to predict disturbance forces based on relative states with nearby quadrotors. During flight, the sequence of disturbance predictions is employed by the controller to mitigate the downwash effect.
By predicting the downwash effect using a neural network, the quadrotor can achieve more accurate trajectory-tracking performance during vertically-aligned flight.

\subsection{Notation and Coordinate Systems}

\setlength{\textfloatsep}{8pt plus 1.0pt minus 2.0pt}
\begin{figure}[t]
    \centerline{\includegraphics[trim=0 0 0 0,clip,width=3.0in]{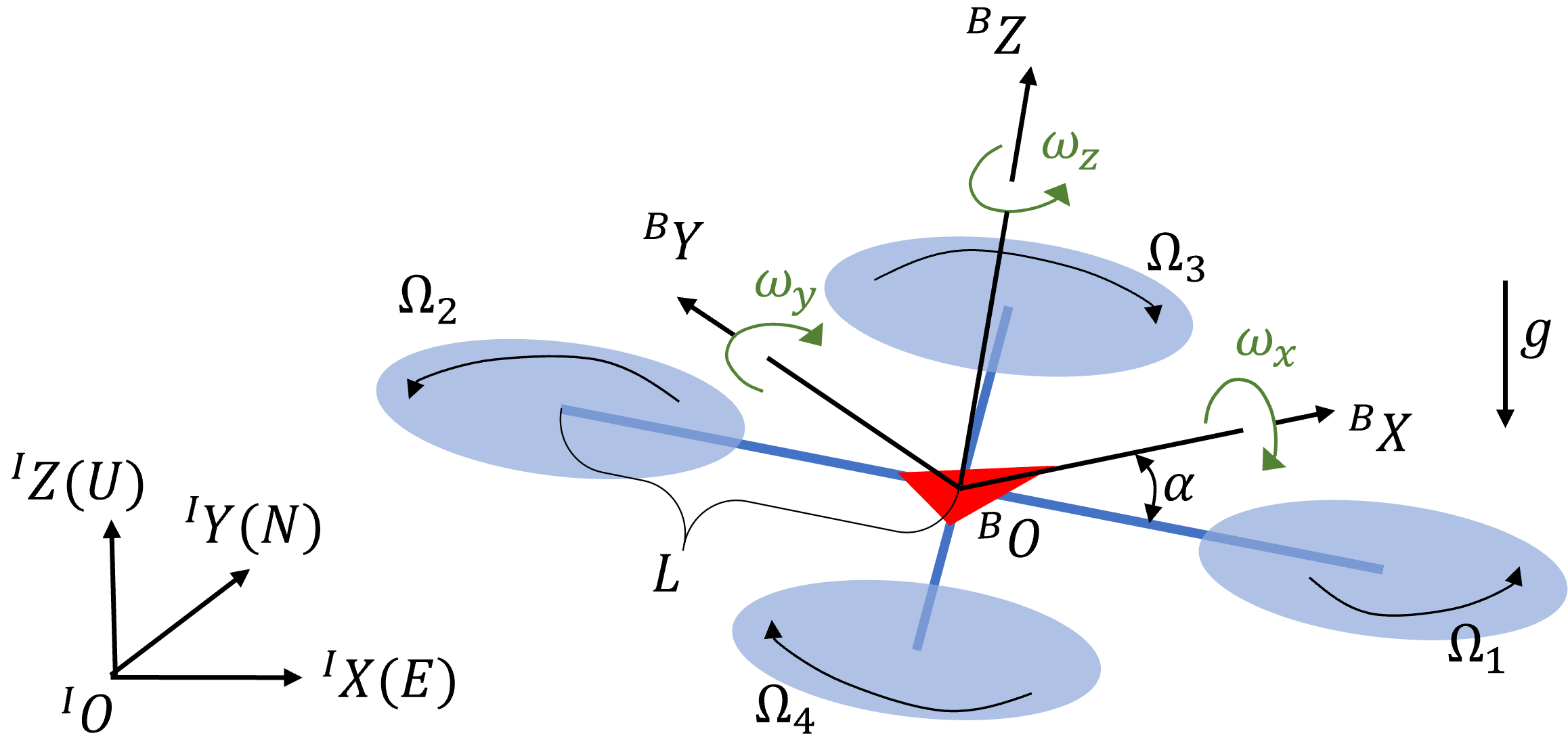}} 
    \vspace*{-1mm}
    \caption{Diagram of a quadrotor model with the ENU (X East, Y North, Z Up) inertial frame and FLU (X Forward, Y Left, Z Up) body frame.}
    \label{fig:coordinate}
\end{figure}


We denote scalars in lowercase $x \in \mathbb{R}$, vectors in bold lowercase $\boldsymbol{x} \in \mathbb{R}^n$, and matrices in bold uppercase $\boldsymbol{X} \in \mathbb{R}^{n \times m}$. We use $[\cdot]$ to denote arrays and $(\cdot)$ to denote functions. We use $\hat{\cdot}$ to denote estimated values. The coordinate systems, depicted in Fig. \ref{fig:coordinate}, contain the world inertial frame $\mathcal{I}$, the body frame $\mathcal{B}$, as well as the propeller numbering convention. 
The vector in the frame $\mathcal{I}$ is denoted as $^{I}\boldsymbol{p}$, and the rotation from $\mathcal{B}$ to $\mathcal{I}$ is denoted as $^{I}_{B}\boldsymbol{R}$ (rotation matrix)  or $^{I}_{B}\boldsymbol{q}$ (attitude quaternion).
We use the ENU inertial frame and FLU body frame to ensure compatibility with MAVROS, a toolkit for communication with autopilots like PX4 \cite{meier_px4_2015}. 

We use $\boldsymbol{q}=[q_w, q_x, q_y, q_z]^T \in \mathbb{H}$ to denote the attitude quaternion in \textit{Hamilton-convention} \cite{sommer_why_2018}, $\boldsymbol{q}^*=[q_w, -q_x, -q_y, -q_z]^T$ to denote the quaternion conjugation, and $\circ$ to denote the quaternion multiplication operator. The attitude quaternion is a unit quaternion ($\left \| \boldsymbol{q} \right\|$=1), and thus its inverse $\boldsymbol{q}^{-1}$ is the same as $\boldsymbol{q}^*$. We use $\mathcal{V}(\cdot)$ to represent the vector part of the quaternion $\mathcal{V}(\boldsymbol{q}) := [{q}_x, {q}_y, {q}_z]^T, \mathbb{H}\rightarrow \mathbb{R}^3$, and $\mathcal{V^*}(\cdot)$ to denote the reverse mapping from a position point $\mathcal{V^*}(\boldsymbol{p}):=[0, \boldsymbol{p}]^T, \mathbb{R}^3\rightarrow \mathbb{H}$. Then full SE3 transformations from $\mathcal{B}$ to $\mathcal{I}$ can be represented as ${^{I}\boldsymbol{p}} = \mathcal{V}({^{I}_{B}\boldsymbol{q}} \circ \mathcal{V^*}({^{B}\boldsymbol{p}}) \circ ^{I}_{B} \boldsymbol{q}^*) + {^{I}\boldsymbol{p}_{Bo}} = {^{I}_{B}\boldsymbol{R}(\boldsymbol{q})} {^{B}\boldsymbol{p}} + {^{I}\boldsymbol{p}_{Bo}}$, where $^{I}\boldsymbol{p}_{Bo}$ is the position of $\mathcal{B}$ frame's origin in the $\mathcal{I}$ frame, and $\boldsymbol{R}(\boldsymbol{q})$ is the rotation matrix from a quaternion following:
$$
\begin{aligned}
\boldsymbol{R}=
\left[\begin{array}{lll}
1-2 q_y^2-2 q_z^2 & 2 q_x q_y-2 q_w q_z & 2 q_x q_z+2 q_w q_y \\
2 q_x q_y+2 q_w q_z & 1-2 q_x^2-2 q_z^2 & 2 q_y q_z-2 q_w q_x \\
2 q_x q_z-2 q_w q_y & 2 q_y q_z+2 q_w q_x & 1-2 q_x^2-2 q_y^2
\end{array}\right].
\end{aligned}
$$

\subsection{Nominal Quadrotor Model}

We assume that the origin of the body frame $\mathcal{B}$ is at the center of mass, and four rotors are all placed in the $\mathcal{B}$ frame's XY-plane. Established from 6-DoF rigid body dynamics, the quadrotor model is written as follows \cite{sun_comparative_2022}
\begin{align}
{^I\dot{\boldsymbol{p}}} &= {^I\boldsymbol{v}}, \label{eq:qd1} \\[5pt]
{^I\dot{\boldsymbol{v}}} &= \left({^{I}_{B}\boldsymbol{R}(\boldsymbol{q})} \cdot {^B\boldsymbol{f}_u} + {^I\boldsymbol{f}_d}\right)/ {m} + {^I\boldsymbol{{g}}}, \label{eq:qd2}\\[5pt]
{^I_B\dot{\boldsymbol{q}}} &= 1/2 \cdot {^I_B\boldsymbol{q}} \circ \mathcal{V^*}(^B\boldsymbol{\omega}), \label{eq:qd3}\\[5pt] 
{^B\dot{\boldsymbol{\omega}}} &=\boldsymbol{{I}}^{-1} \cdot\left(-{^B \boldsymbol{\omega}} \times\left({\boldsymbol{{I}}} \cdot {^B \boldsymbol{\omega}}\right)+{^B\boldsymbol{\tau}_u} + {^B\boldsymbol{\tau}_d}\right), \label{eq:qd4}
\end{align}
where $m$ is the mass, ${^I\boldsymbol{{g}}}=[0,0,{-g}]^T$ is  the gravity vector, $\boldsymbol{{I}}=\texttt{diag} ({I}_{xx}, {I}_{yy}, {I}_{zz})$ is the inertia matrix assuming that quadrotors exhibit symmetry across all three axes, ${^B\boldsymbol{f}_u}$ and ${^B\boldsymbol{\tau}_u}$ are the force and torque caused by rotors, ${^I\boldsymbol{f}_d}$ and ${^B \boldsymbol{\tau}_d}$ are the force and torque caused by disturbances, and ${^B\boldsymbol{\omega}}=\left[\omega_x,\omega_y,\omega_z \right]^T$ is the angular rate expressed in the $\mathcal{B}$ frame.

The thrust generated by rotors is assumed to be vertical to the $\mathcal{B}$ frame's XY-plane, and we therefore obtain ${^B\boldsymbol{f}_u}=\left[0,0,f_c \right]^T$ and ${^B\boldsymbol{\tau}_u}=\left[\tau_x, \tau_y, \tau_z \right]^T$, where $f_c$ is the collective force of four rotors. We use a quadratic fit to model the thrust and torque for each propeller:
\begin{equation}
    f_i={k}_{t} \cdot \Omega^2, \quad \tau_i={k}_{q} \cdot \Omega^2,
    \label{eq:2}
\end{equation}
where ${k}_{t}$ and ${k}_{q}$ are the thrust coefficient and torque coefficient, respectively, as well as $\Omega$ represents motor speed in kRPM.
Then $[f_c, \tau_x, \tau_y, \tau_z]^T$ and the thrust of each rotor $f_i$ is connected by
\begin{equation}
    \left[f_c, \tau_x, \tau_y, \tau_z \right]^T = \boldsymbol{G} \cdot \left[f_1, f_2, f_3, f_4 \right]^T,
    \label{eq:3}
\end{equation}
in which the control allocation matrix $\boldsymbol{G}$ is
\begin{equation}
\boldsymbol{G}=\left[\begin{array}{cccc}
1 & 1 & 1 & 1 \\
-{L} \sin \mathbf{\alpha} & {L} \sin \mathbf{\alpha} & {L} \sin \mathbf{\alpha} & -{L} \sin \mathbf{\alpha} \\
-{L} \cos \mathbf{\alpha} & {L} \cos \mathbf{\alpha} & -{L} \cos \mathbf{\alpha} & {L} \cos \mathbf{\alpha} \\
-{k}_{q} / {k}_{t} & -{k}_{q} / {k}_{t} & {k}_{q} / {k}_{t} & {k}_{q} / {k}_{t}
\end{array}\right],
\end{equation}
where ${L}$ and ${\alpha}$ are the geometric parameters as in Fig. \ref{fig:coordinate}.

The quadrotor's nominal model established above is utilized for designing both the disturbance observer and controller. The disturbance ${^B \boldsymbol{\tau}_d}$ is compensated by a high-frequency body-rate controller within the autopilot, and the ${^I\boldsymbol{f}_d}$ is estimated in the subsequent section.

\subsection{Neural Network Observer for Downwash Effect}

This part introduces a neural network observer to model the disturbance between quadrotors in close-proximity flight.

\subsubsection{Neural Network Disturbance Observer}


Considering the high nonlinearity of airflow, we employ a Multi-Layer Perceptron (MLP) to estimate the disturbances. A trained MLP can be viewed as a mapping function $\boldsymbol{y}=f(\boldsymbol{x}; \boldsymbol{\theta}):\mathbb{R}^i\rightarrow\mathbb{R}^o$ from the input $\boldsymbol{x}$ to the output $\boldsymbol{y}$, where $\boldsymbol{\theta}:=\left\{\boldsymbol{W}^1, b^1,\cdots,\boldsymbol{W}^{{H}+1}, b^{H+1}\right\}$ represent the weight and bias parameters, and ${H}$ is the number of hidden layers. By choosing the element-wise ReLU  $\phi(\boldsymbol{x})=\max(0, \boldsymbol{x})$ as the activation function, the MLP network can be written as
\begin{equation}
f\left(\boldsymbol{x}; \boldsymbol{\theta}\right)=\boldsymbol{W}^{H+1} \cdot \phi\left(\cdots\phi\left(\boldsymbol{W}^1\boldsymbol{x}+b^1\right)\cdots\right)+b^{H+1}.
\label{eq:5}
\end{equation}

When applying the network to model the downwash effect, the input variables encompass the relative position and velocity of the ego quadrotor and the other one as $\boldsymbol{x} = [{^I\boldsymbol{p}_{rela}}, {^I\boldsymbol{v}_{rela}}]^T$, while the outputs encompass the disturbances as $\boldsymbol{y}= {}^I\hat{\boldsymbol{f}}_d$.


\subsubsection{Spectral Normalization}

The training set we collected is impossible to cover the entire state space, and thus the output of the network in those data-uncovered states is critical to flight safety. 
Spectral normalization has been demonstrated in recent papers \cite{shi_neural_2019}, \cite{miyato_spectral_2018} that it can enhance the robustness and generalization of neural networks and hence be adopted.

The Lipschitz constant of a function $\|f\|_\text{Lip}$ is defined as the smallest value such that
\begin{equation}
\forall \boldsymbol{x},\boldsymbol{x'}: \left\| f(\boldsymbol{x}) - f(\boldsymbol{x'}) \right\|_2 / \left\| \boldsymbol{x} - \boldsymbol{x'} \right\|_2 \leq \|f\|_\text{Lip}.
\end{equation}
Let $l(\boldsymbol{x})=\boldsymbol{Wx}+\boldsymbol{b}$, and then we can get the Lipschitz norm of one layer in (\ref{eq:5}):
\begin{equation}
    \frac{\left\|\boldsymbol{Wx}+\boldsymbol{b}-\boldsymbol{Wx'}-\boldsymbol{b} \right\|_2}{\left\|\boldsymbol{x} - \boldsymbol{x'}\right\|_2} \leq \left\| \boldsymbol{W} \right\|_2 = \sigma(\boldsymbol{W}) = \|l\|_\text{Lip},
\end{equation}
where $\sigma(\cdot)$ represents spectral norm, i.e., the maximum singular value.
Leveraging the Lipschitz constant inequality of composite functions $\left\|g_1 \circ g_2\right\|_{\text {Lip}} \leq\left\|g_1\right\|_{\text {Lip}} \cdot \left\|g_2\right\|_{\text {Lip}}$, and the fact that the Lipschitz constant of ReLU function $\left\|\phi(\boldsymbol{x})\right\|_\text{Lip}=1$, we can obtain the bound of the MLP (\ref{eq:5}):
\begin{equation}
    \left\|f(\boldsymbol{x})\right\|_\text{Lip} \leq \sigma(\boldsymbol{W}^{H+1}) \cdot 1 \cdots \sigma(\boldsymbol{W})
    =\prod_{l=1}^{{H}+1} \sigma\left(\boldsymbol{W}^l\right).
\end{equation}

In training phase, if every $\boldsymbol{W}^l$ is normalized by its spectral norm $\sigma(\cdot)$ and the scale ratio $\gamma$ at each training epoch
\begin{equation}
    \overline{\boldsymbol{W}}^l:=\gamma \cdot \boldsymbol{W}^l / \sigma\left(\boldsymbol{W}^l\right),
\end{equation}
then the Lipschitz norm of the network can be bounded as
\begin{equation}
    \left\|f(\boldsymbol{x};\overline{\boldsymbol{W}}^l)\right\|_\text{Lip} \leq \gamma^{{H}+1}.
\end{equation}

Spectral normalization effectively limits the change rate of the network's output and leads to a more uniform output. This uniformity is further substantiated by our subsequent experiments.

\subsubsection{Data Acquisition}
Collecting the disturbance force is vital for training the network. Assuming that the physical model obtained through parameter identification is accurate, the disturbance force can be computed by subtracting the nominal force from the real one.

The nominal resultant force ${^I\boldsymbol{f}_n}$ can be calculated as follows assuming that the motor speed $\Omega_i$ is measured during the training phase:
\begin{equation}
        {^I\boldsymbol{f}_n} = {^{I}_{B}\boldsymbol{R}(\boldsymbol{q})} \cdot {^B\boldsymbol{f}_u} + {m} \cdot {^I\boldsymbol{{g}}},
\end{equation}
where $^B\boldsymbol{f}_u$ comes from (\ref{eq:2}) and (\ref{eq:3}). In addition, the real resultant force ${^I\boldsymbol{f}}$ can be acquired using odometry. The odometry module of an autopilot offers velocity estimates, and its derivative $^I \dot{\hat{\boldsymbol{v}}}$ is numerically calculated using Tustin's method. Subsequently, the real resultant force on the body is obtained through classical mechanics:
\begin{equation}
    {^I\boldsymbol{f}} ={m}\cdot{ }^I \dot{\hat{\boldsymbol{v}}}.
\end{equation}

Finally, the disturbance $^I\boldsymbol{f}_d$ is determined by
\begin{equation}
    ^I\boldsymbol{f}_d = {^I\boldsymbol{f}} - {^I\boldsymbol{f}_n}.
\end{equation}
Now the inputs $\left[{^I\boldsymbol{p}_{rela}}, {^I\boldsymbol{v}_{rela}}\right]$ and outputs ${^I\boldsymbol{f}_d}$ have been constructed for training the neural network.

\textit{\textbf{Remark}:} The resultant force can also be estimated directly using the inertial measurement unit (IMU), but the noise level is unacceptable. In addition, predictions for a sequence of states can be computed collectively in a single batch when utilizing the network in real flight.


\subsection{Nonlinear MPC with Network Disturbance Prediction}

This subsection first introduces the generation of the control target, i.e., the reference trajectory. Subsequently, the detailed presentation of the proposed NMPC-based control approach follows.

\subsubsection{Reference Trajectory Generation}

Trajectory generation involves the creation of a smooth, dynamically feasible, and time-indexed curve that traverses a set of points, including a start point, multiple predefined waypoints, and an end point. Leveraging the inherent mathematical property of \textit{differential flatness} in quadrotors, the process of trajectory generation can be reformulated into a polynomial optimization problem for the four flat outputs $[x, y, z, \psi]$ corresponding to 3D position and yaw angle. To achieve this, we implement the \textit{minimum snap} algorithm as outlined in \cite{mellinger_minimum_2011}, except that the time allocation among points is accomplished by straightforwardly dividing the relative distance by the user-defined average velocity.

The generated trajectory comprises a collection of parametric equations, which requires a trajectory server to discretize the curve. This server is also responsible for selecting the future reference states that align with the prediction horizon of NMPC, and then publishing these references. The publication frequency is set to be the same as the control frequency.

\subsubsection{Nonlinear MPC}

During close-proximity flight, we assume that each quadrotor employs an NMPC controller. This enables them to share predictions among themselves, ensuring the possibility of future disturbance predictions. Under above assumption,  we select Nonlinear MPC for close-proximity flight due to two primary reasons. First, the ego quadrotor can prepare for the downwash effect in advance through the prediction trajectory of another quadrotor. Second, the potential saturation of the thrust command, which arises as one quadrotor experiences the downwash airflow, can be powerfully handled by NMPC.

Given the reference trajectory, the cost function is defined as the accumulated error between predicted and reference states over the time horizon. Then a constrained nonlinear optimization problem is formulated as
\begin{equation}
\label{eq:cost function}
    \min\limits_{\boldsymbol{u}_k}\left( \overline{\boldsymbol{x}}^T_N \boldsymbol{{Q}}_N\overline{\boldsymbol{x}}_N + \sum\limits_{k=0}^{N-1}\left(\overline{\boldsymbol{x}}^T_k \boldsymbol{{Q}}\overline{\boldsymbol{x}}_k + \overline{\boldsymbol{u}}^T_k \boldsymbol{{R}}\overline{\boldsymbol{u}}_k\right)\right)
\end{equation}
with constraints of dynamics, initial values, and control inputs:
\begin{equation}
\begin{aligned}
    \boldsymbol{x}_{k+1} &= f\left(\boldsymbol{x}_{k}, \boldsymbol{u}_{k}\right),\\ \boldsymbol{x}_0 &=\boldsymbol{x}_{\mathrm{init}},\\ \boldsymbol{u}_k&\in\left[\boldsymbol{u}_{\mathrm{min}}, \boldsymbol{u}_{\mathrm{max}}\right],
\end{aligned}
\end{equation}
where the symbol $\overline{(\cdot)}=(\cdot) - (\cdot)_r$ denotes the error w.r.t. the reference, $\boldsymbol{x}_{\mathrm{init}}$ is the current quadrotor state, the diagonal matrices $\boldsymbol{{Q}}_N, \boldsymbol{{Q}}, \boldsymbol{{R}}$ represent weights for terminal cost, state cost, and control energy cost, respectively, and $f(\cdot)$ is the quadrotor nominal model (\ref{eq:qd1}-\ref{eq:qd3}) discretized by the 4th-order \textit{Runge-Kutta} method. Specifically, the state $\boldsymbol{x}_k$ equals to $[{^I\boldsymbol{p}}_k,{^I\boldsymbol{v}}_k,{^I\boldsymbol{q}}_k]^T$, and the control input $\boldsymbol{u}_k$ equals to $[f_c, {^B\boldsymbol{\omega}}]^T$. Note that (\ref{eq:cost function}) is a nonlinear least squares cost due to the nonlinearity introduced by quaternions. If the quaternion error is denoted as
${\boldsymbol{q}}_e=\boldsymbol{q} \circ \boldsymbol{q}_r^{-1}$, considering the fact that only three variables in a quaternion are independent, we can write the quaternion term in the cost function as
\begin{equation}
    \overline{\boldsymbol{q}}^T_k \boldsymbol{{Q}}_q\overline{\boldsymbol{q}}_k
    =\left\|\text{sgn}(q_{ew})\cdot\mathcal{V}({\boldsymbol{q}}_e)\right\|_{\boldsymbol{Q}}^2
    =\mathcal{V}({\boldsymbol{q}}_e)^T \boldsymbol{{Q}}_q \mathcal{V}({\boldsymbol{q}}_e),
\end{equation}
where $\text{sgn}(\cdot)$ denotes the sign function. The sign term, which can avoid the unwinding phenomenon in quaternion-based control \cite{mayhew_quaternion-based_2011}, is able to be eliminated in the quadratic cost.

\setlength{\textfloatsep}{8pt plus 1.0pt minus 2.0pt}
\begin{figure}[t]
    \centerline{\includegraphics[trim=0 0 0 0,clip,width=\linewidth]{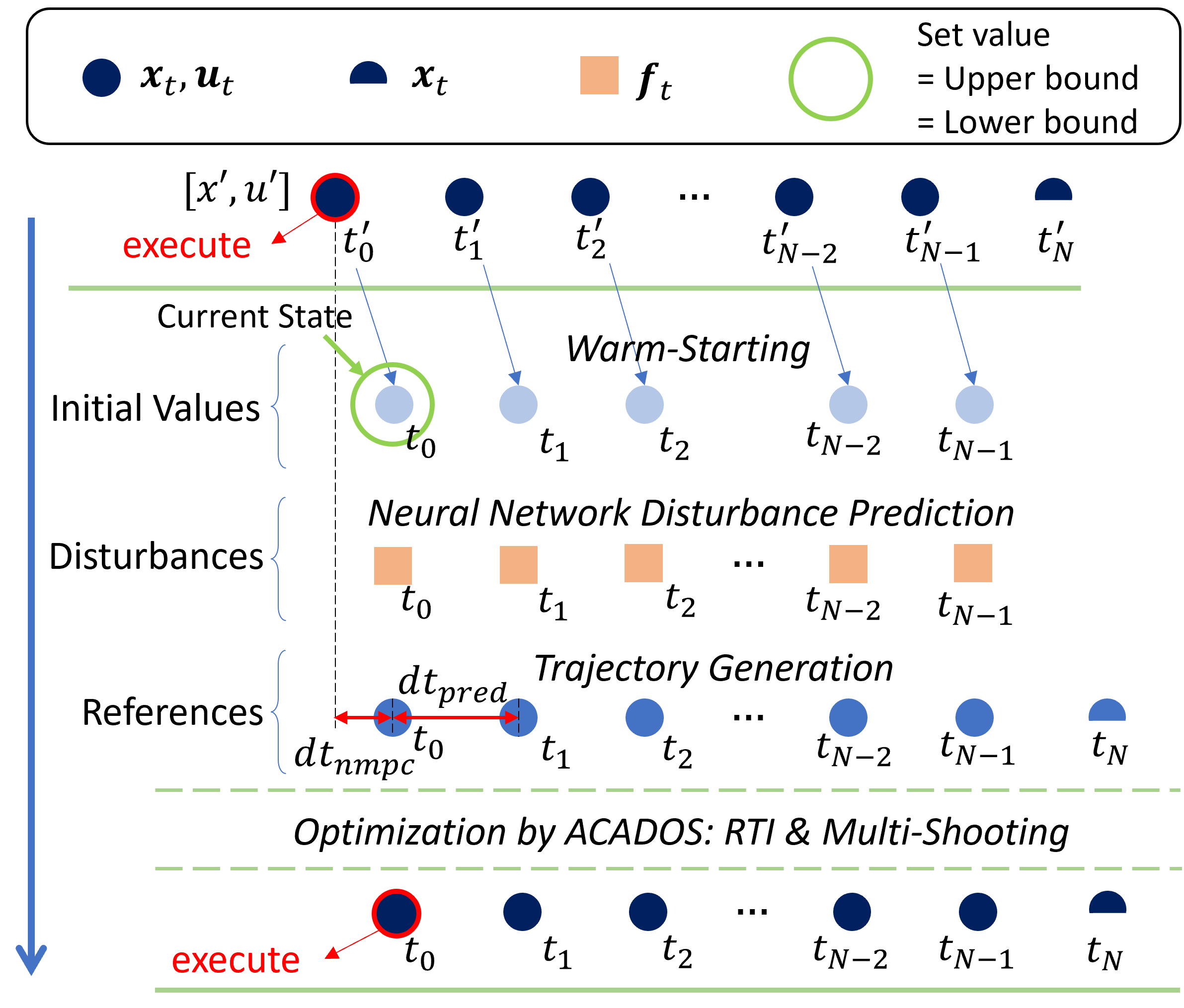}} 
    \caption{A diagram illustrating the NDP-NMPC algorithm. From top to bottom, the initial value of each state point in this iteration is derived from the result of the previous iteration, a concept known as \textit{warm-starting}. Then the first state is constrained to match the current state, introducing a feedback mechanism into NMPC. Next, the disturbances are predicted by a neural network and used as parameters for the optimizer, as well as the reference states are provided by the trajectory server. Finally, these data points are transmitted to ACADOS \cite{verschueren_acadosmodular_2022} for optimization.}
    \label{fig:nmpc}
\end{figure}

Subsequently, \textit{warm-starting}, \textit{real-time iteration (RTI)}, and \textit{multi-shooting} techniques \cite{verschueren_acadosmodular_2022} are applied to accelerate the NMPC computation, which is illustrated in Fig. \ref{fig:nmpc}. Finally, the control command is extracted from the optimized result sequence:
\begin{equation}
    \boldsymbol{u}_\mathrm{NDP-NMPC}=\boldsymbol{u}^*_0=\left[f'_c, \omega'_x, \omega'_y, \omega'_z\right].
\end{equation}

Note that the thrust command $f'_c$ needs to be normalized into $[0,1]$ to align with the MAVROS toolkit. This normalization can be implemented either by a Kalman Filter to estimate the hover throttle, similar to the approach used in the PX4 autopilot \cite{grob_quaternion_2016}, or through a calibration mapping that relates thrust to throttle.

\section{Experiments}

This section introduces the experimental setup and analyzes the results. We begin by presenting the system identification of our hardware platform. Following that, we discuss the data collection related to the downwash effect. Next, we describe an open-loop experiment conducted for disturbance prediction. Finally, we perform a closed-loop trajectory tracking experiment to validate the control performance.

\subsection{Parameter Identification for Quadrotors}

\begin{figure}[t]
    \centering
    \subfloat[Flight platform]{
        \includegraphics[trim=0 0 0 0,clip,height=1.35in]
        {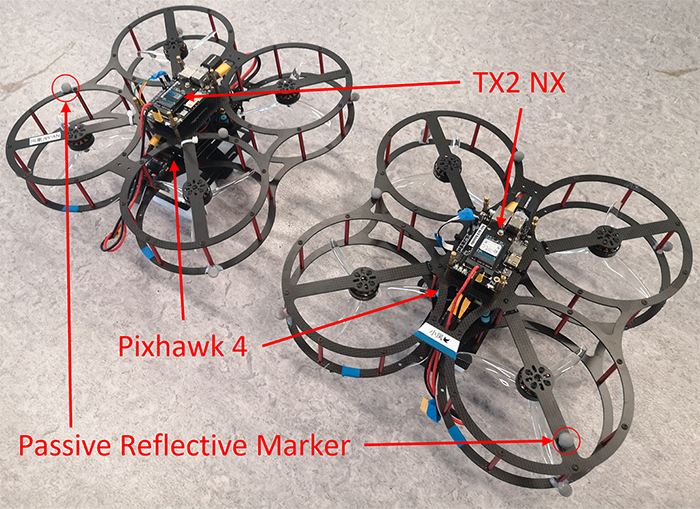}
        \label{fig:quad_real}
    }
    \subfloat[Parameter identification]{
        \includegraphics[height=1.35in]{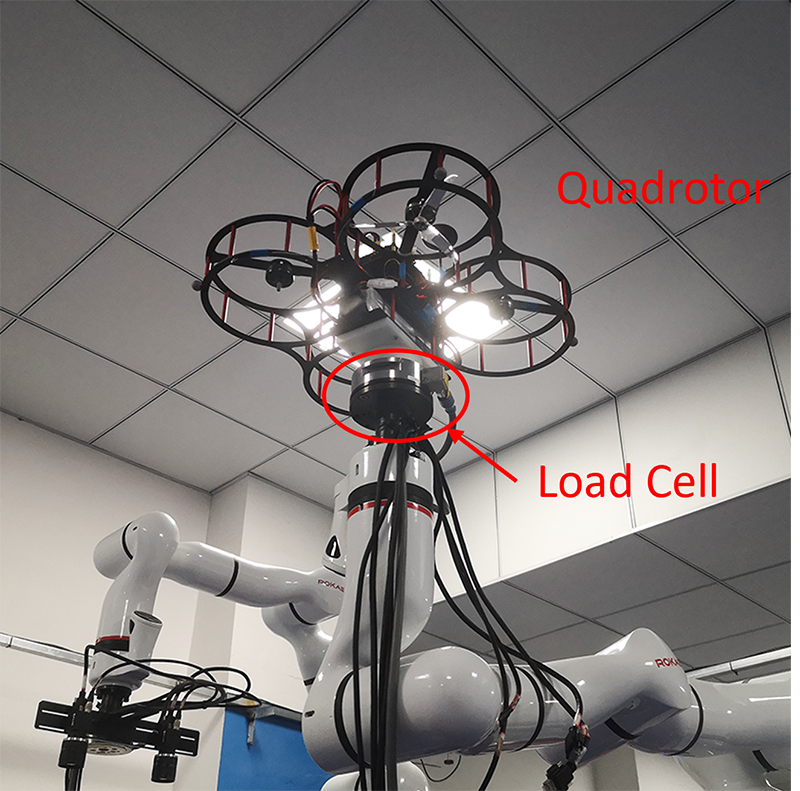}
        \label{fig:sysid}
    }
    \caption{(a) Two self-made quadrotors with onboard computing resources. (b) The quadrotor is identified for rotor parameters and inertial parameters.}
    \label{fig:hardware}
\end{figure}

We constructed our flight platform as depicted in Fig. \ref{fig:quad_real}. It was noteworthy that the selected Electronic Speed Controllers (ESCs) supported rotor speed measurement through the DShot protocol\footnote{\href{https://docs.px4.io/main/en/peripherals/dshot.html}{https://docs.px4.io/main/en/peripherals/dshot.html}}. Then, we leveraged a load cell as illustrated in Fig. \ref{fig:sysid} to identify the rotor parameters. Specifically, a 3D-printed connector was used to position the quadrotor on the load cell, and square wave signals were applied to drive a pair of propellers diagonally. This setup took into account the airflow effect of the body on the rotors. Assuming that the quadrotor is symmetrical in all three axes, we measured its inertial matrix using the bifilar-pendulum method \cite{jardin_optimized_2009}. The identified parameters listed in Table \ref{tab:qd_params} were employed for both PX4 SITL simulation and control.



\setlength{\dbltextfloatsep}{8pt plus 1.0pt minus 2.0pt}


\begin{table}[t]
\centering
\caption{Identified Parameters}
\begin{tabular}{ ccl } 
 \toprule
  \textbf{Parameter(s)} & \textbf{Value(s)} & \textbf{Unit} \\
 \midrule
 ${L}$ & 0.1372 & m \\ 
 ${\alpha}$ & 45 & deg \\ 
 ${m}$ & 1.5344 & kg \\ 
 ${g}$ & 9.81 & $\text{m}/\text{s}^2$ \\ 
 ${I}_{xx}$ & 0.0094 & $\text{kg}\cdot\text{m}^2$ \\
 ${I}_{yy}$ & 0.0134 & $\text{kg}\cdot\text{m}^2$ \\
 ${I}_{zz}$ & 0.0145 & $\text{kg}\cdot\text{m}^2$ \\
 ${k}_{q}$ & 3.7611 E-4 & $\text{N} \cdot  \text{m}/\text{kRPM}^2$ \\
 ${k}_{t}$ & 2.8158 E-2 & $\text{N}/\text{kRPM}^2$ \\
 $[\Omega_\mathrm{min}, \Omega_\mathrm{max}]$ & [2.6, 24.0] & kRPM \\
 thrust/weight & 4.3100 & — \\
 flight time & 705 & s \\
 \bottomrule
\end{tabular}
\label{tab:qd_params}
\end{table}

\subsection{Data Collection, Training, and Testing}

To collect data points under the downwash effect, two quadrotors were operated by pilots to fly over each other. The lower quadrotor remained stationary while the upper quadrotor was moved to increase the likelihood of overlap. The height of the moving quadrotor was varied to diversify the data collection. Data from both quadrotors, including their states and estimated disturbances along with timestamps, were captured at a rate of 100Hz using the ROS tool \texttt{rosbag}. A total of 570-second data were collected for both quadrotors, and this data size can be doubled due to the similarity of drones. Then the data were bias eliminated by subtracting the average disturbance force in the hover state and time-aligned. Finally, the collected data were shuffled with a ratio of 0.75 for training and 0.25 for testing.


    

In training, the network was implemented in PyTorch with the parameters detailed in Table \ref{tab:nn_para}. As previously mentioned, the input consisted of the 3-axis relative position and velocity, while the output included the 3-axis disturbance force. The network was trained with various spectral normalization ratios $\gamma$ to determine the optimal value.

\setlength{\textfloatsep}{8pt plus 1.0pt minus 2.0pt}
\begin{figure}[t]
    \centerline{\includegraphics[trim=0.0cm 4.2cm 0.0cm 4.2cm,clip,width=\linewidth]{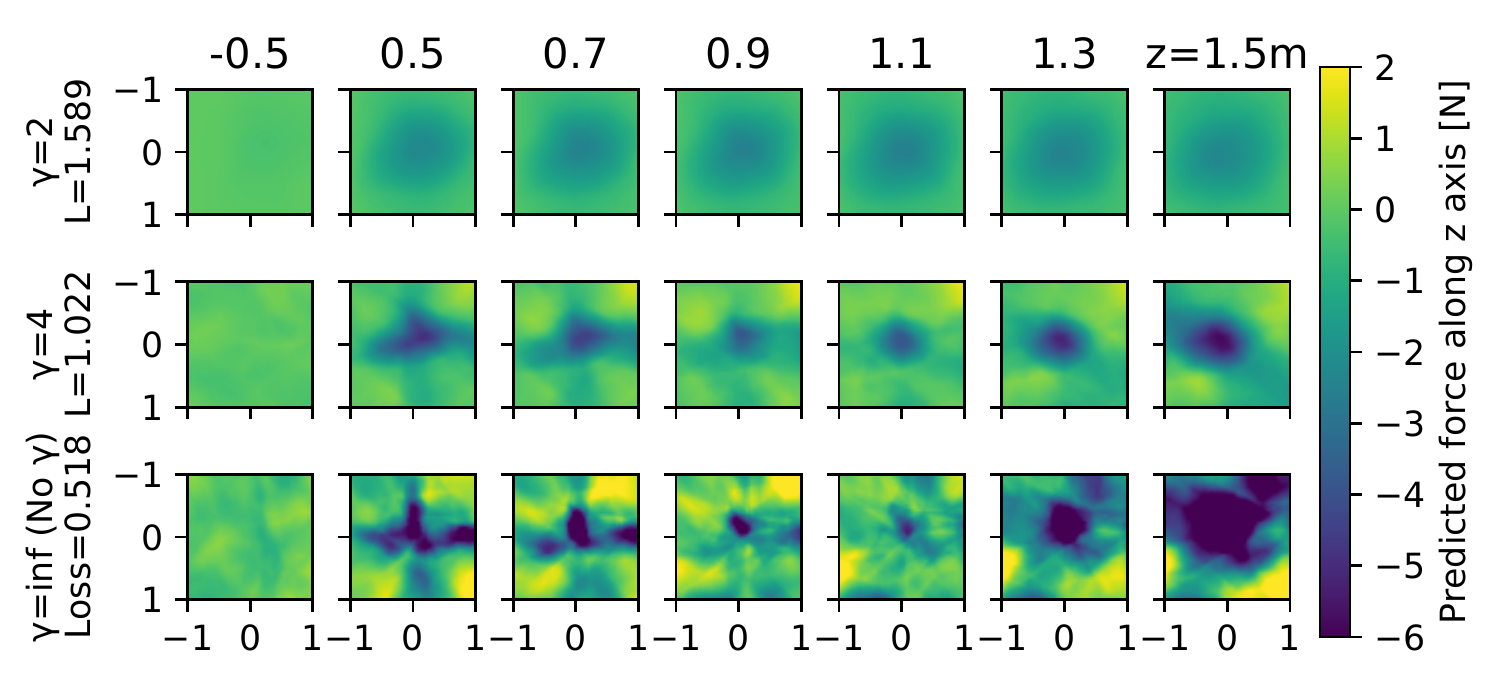}} 
    \caption{Disturbance predictions of the neural network under different spectral normalization ratios $\gamma$. Each square represents the predicted Z force in a 1m$\times$1m X-Y area. Given zero relative velocity, we change the $\gamma$ to observe the output at different heights. The relative heights of the other quadrotor to the ego one are listed from left to right, where the negative number indicates flying above. The disturbance predictions for the X and Y axes are not displayed, as they are considered negligible when compared to the Z force.}
    \label{fig:pred}
\end{figure}

\captionsetup{position=top}
\begin{table}[b]
  \centering
  \caption{Settings for Training the Network}
\begin{tabularx}{\linewidth}{ llll}  
 \toprule
 \multicolumn{1}{l|}{\textbf{Setting}}  & \textbf{Layers} & \textbf{Initialization} & \textbf{Activation Function}   \\
 \midrule
 \multicolumn{1}{l|}{\textbf{Value}} & 6-128-64-128-3 & normal & ReLU   \\
 \midrule
 \textbf{Optimizer} & \textbf{Epoch} & \textbf{Learning Rate} & \textbf{Loss Function} \\
 \midrule
 Adam & 20,000 & 1e-4 & Mean Squared Error \\
    
 \bottomrule
\end{tabularx}%
\label{tab:nn_para}
\end{table}

The spectral normalization ratio $\gamma$ is critical to the balance of accuracy and robustness. We recorded the losses for different values of $\gamma$ and plotted the predictions at different heights as Fig. \ref{fig:pred} to evaluate performance in cases where no data are available.

From the figure, we can observe that a high value of $\gamma$ (the third row) describes the disturbances well, while a low value of $\gamma$ (the first row) is too conservative. In other words, the stronger the influence of spectral normalization, the lower the accuracy in fitting data. However, this low $\gamma$ value results in safer predictions for points where no data have been collected. For the third row without spectral normalization, the predicted force suddenly increases at 1.3m due to the lack of data. Nevertheless, the first and second rows with Low $\gamma$ mitigate this trend and guarantee flight safety. In conclusion, we decide to choose $\gamma=4$ as the balance between data fitting and safety.

\subsection{Trajectory Tracking under Downwash Effect}

In this section, we aim to close the loop and verify the feasibility of the proposed method considering hardware limitations such as system latency and constrained computational resources for embedded platforms. The entire system was developed in Python and leveraged Robot Operating System (ROS) for communication. The system operated on both an onboard TX2 NX computer and a desktop PC. Specifically, the trajectory server, NDP-NMPC controller, and hover throttle estimator ran on the TX2 NX, while the PC was responsible for trajectory generation, broadcasting motion capture (MoCap) data, and serving as the ROS master node. The NMPC method was implemented using ACADOS with parameters listed in Table \ref{tab:control_params}. We assigned high weights to positions in order to minimize tracking errors.

 

\begin{table}[b]
\centering
\caption{Control Parameters}
\begin{tabular*}{\linewidth}{@{\extracolsep{\fill}} c c|c c|c c }
 \toprule
  \textbf{Parameter} & \textbf{Value} & \textbf{Parameter} & \textbf{Value} & \textbf{Parameter} & \textbf{Value} \\
\midrule
$N$ & 20 & $dt_{\rm nmpc}$ & 1/60s & $dt_{\rm pred}$ & 0.1s \\
$Q_{p,\rm xy}$ & 300 & $Q_{p,\rm z}$ & 400 &
$Q_{\rm v,xyz}$ & 1 \\
$Q_{q,\rm xyz}$ & 0.1 & $R_{\omega,\rm xyz}$ & 10  & $R_{f_c}$ & 10  \\
 
 \bottomrule
\end{tabular*}
\label{tab:control_params}
\end{table}

\begin{figure}[t]
    \centering
    \centerline{\includegraphics[trim=10 10 40 65,clip,width=\linewidth]{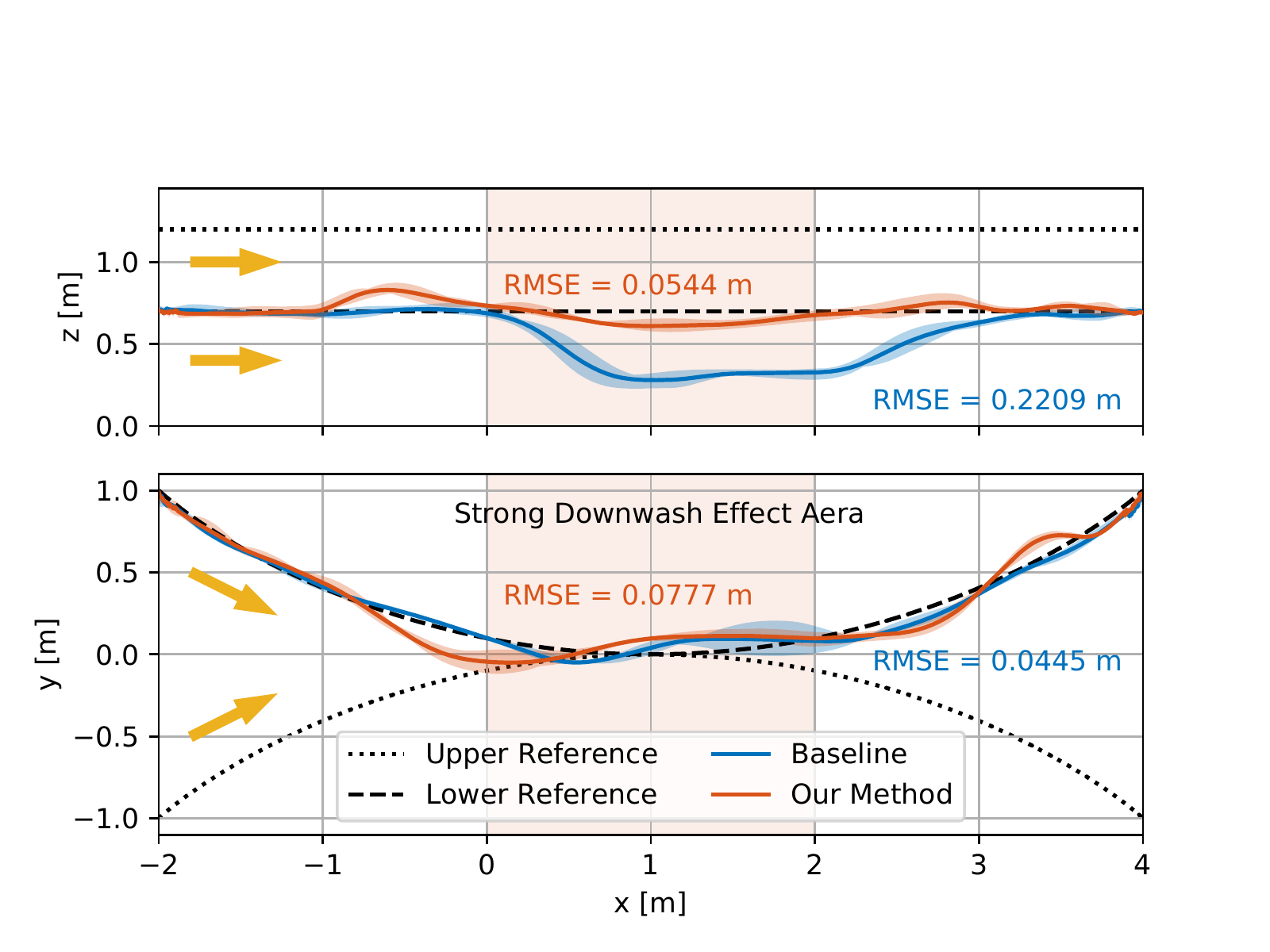}} 
    \caption{Comparison of closed-loop tracking results between the baseline \cite{falanga_pampc_2018} and the proposed method. Each curve represents the average of three rounds, with the shaded area indicating the range of values. The shaded middle region highlights the zone with a strong downwash effect.}
    \label{fig:close_result}
\end{figure}

We conducted quadrotor flights within a room measuring $7\times4\times3$ and used an OptiTrack MoCap system for localization. The flight trajectories were carefully designed to ensure an overlapping area. During the experiments, two quadrotors initiated from the same side but at different altitudes, and then executed back-and-forth movements. To evaluate the performance of closed-loop tracking, we conducted multiple runs, and the resulting trajectories are illustrated in Fig. \ref{fig:close_result}.  We established the NMPC method without disturbance prediction \cite{falanga_pampc_2018} as the baseline.
 


\setlength{\dbltextfloatsep}{8pt plus 1.0pt minus 2.0pt}

From Fig. \ref{fig:close_result}, the baseline quadrotor is severely affected and thus deviates from the reference. In contrast, NDP-NMPC significantly mitigates the impact of disturbances, leading to a noteworthy 75.37\% reduction in the tracking Root-Mean-Square Error (RMSE). This reduction emphasizes the positive impact of NDP-NMPC on close-proximity flight.
Additionally, we observe that the lower quadrotor suddenly jumps before it enters the downwash area and then is pushed down to the reference height. This phenomenon is caused by the inaccuracy of neural network predictions. In other words, the neural network predicts a disturbance that does not exist in reality and has no corresponding counterbalancing force, which results in an unexpected ascent of the lower quadrotor.
This prediction inaccuracy also introduces fluctuations in the horizontal plane as shown in Fig. \ref{fig:close_result}. Therefore, when attempting to integrate predictions to enhance the system performance, it is critical to carefully weigh the potential implications.


\section{Conclusion}

In this article, we proposed NDP-NMPC, a trajectory tracking method to alleviate the disturbance from downwash airflow in close-proximity flight. This approach utilized a neural network with spectral normalization to predict the disturbances caused by other quadrotors flying above. The network observer was then combined with an NMPC controller. To test the method, we trained a neural network and evaluated the performance of the disturbance prediction. Finally, we executed the algorithm on two quadrotors in real-time for trajectory tracking. We reported that the network was robust, and the proposed approach reduced 75.37\% tracking error in the Z axis.

In the future, the proposed approach should be compared with non-prediction methods to better highlight the advantage of integrating predictions. Additionally, future directions will extend the proposed method to large-scale swarm drones and consider the uncertainty of predictions inside the NMPC workflow.
 

\bibliographystyle{IEEEtranBST2/IEEEtran}
\bibliography{IEEEtranBST2/IEEEabrv, IEEEtranBST2/references}


\end{document}